%% file: main.tex
\documentclass[journal]{IEEEtran}
\pdfoutput=1

\usepackage[a4paper, margin=1in]{geometry}
\usepackage{setspace}
\usepackage{lineno}

\usepackage[numbers]{natbib}

\usepackage{graphicx}
\usepackage{amssymb}
\usepackage{amsmath}
\usepackage{color, colortbl}
\graphicspath{{Figures}}
\DeclareGraphicsExtensions{.pdf,.png}
\usepackage{csquotes}

\usepackage[T1]{fontenc}
\usepackage{algorithm, setspace}
\usepackage[noend]{algpseudocode}
\usepackage{url}
\usepackage{multirow}
\usepackage{float}
\usepackage{xcolor}
\usepackage{hyperref}
\usepackage{tabularx}
\hypersetup{
     colorlinks=true,
     linkcolor=orange,
     filecolor=orange,
     citecolor=orange,      
     urlcolor=orange,
     }

\newcommand{\p}[1]{\smallskip \noindent \textbf{{#1}.}}
\newcommand{\eq}[1]{Equation~(\ref{eq:#1})}
\newcommand{\fig}[1]{Figure~\ref{fig:#1}}

\usepackage{balance}
\usepackage{enumitem}

\ifCLASSINFOpdf
\else
\fi

\begin{document}
\title{Safely and Autonomously Cutting Meat \\ with a Collaborative Robot Arm}

\author{Ryan Wright$^*$,
        Sagar Parekh$^*$,
        Robin White,
        and Dylan P. Losey% <-this % stops a space
\thanks{Sagar Parekh and Ryan Wright contributed equally to this work. Ryan Wright and Robin White are with the Dept. of Animal and Poultry Science, Virginia Tech, Blacksburg, VA, USA. Sagar Parekh and Dylan Losey are with the Dept. of Mechanical Engineering, Virginia Tech, Blacksburg, VA, USA.}% <-this % stops a space
\thanks{Corresponding author can be contacted at: Dylan Losey, Goodwin Hall, 635 Prices Fork Road, Blacksburg, VA 24061, USA. Email: \texttt{losey@vt.edu}}}

\maketitle

\begin{abstract}
Labor shortages in the United States are impacting a number of industries including the meat processing sector. Collaborative technologies that work alongside humans while increasing production abilities may support the industry by enhancing automation and improving job quality. However, existing automation technologies used in the meat industry have limited collaboration potential, low flexibility, and high cost. The objective of this work was to explore the use of a robot arm to collaboratively work alongside a human and complete tasks performed in a meat processing facility. Toward this
objective, we demonstrated proof-of-concept approaches to ensure human safety while exploring the capacity of the robot arm to perform example meat processing tasks. In support of human safety, we developed a knife instrumentation system to detect when the cutting implement comes into contact with meat within the collaborative space. To demonstrate the capability of the system to flexibly conduct a variety of basic meat processing tasks, we developed vision and control protocols to execute slicing, trimming, and cubing of pork loins. We also collected a subjective evaluation of the actions from experts within the U.S. meat processing industry. On average the experts rated the robot's performance as adequate. Moreover, the experts generally preferred the cuts performed in collaboration with a human worker to cuts completed autonomously, highlighting the benefits of robotic technologies that assist human workers rather than replace them. Video demonstrations of our proposed framework can be found here: \url{https://youtu.be/56mdHjjYMVc}

\end{abstract}


% \fi
% \IEEEpeerreviewmaketitle

\input{intro}

\input{method}
\input{results_discussion}

\input{conclusion}

% % Can use something like this to put references on a page
% % by themselves when using endfloat and the captionsoff option.
% \ifCLASSOPTIONcaptionsoff
%   \newpage
% \fi

% \bibliographystyle{elsarticle-harv}
\bibliographystyle{plain}

\bibliography{citations}

\end{document}

%% file: intro.tex
\section{Introduction} \label{sec:intro}

Meat processing is a critical worldwide industry that has made great strides in safety and productivity over the last few decades. However, meat processing facilities across the United States are currently facing labor shortages that negatively impact facility owners, livestock producers, and food consumers. This labor shortage was exacerbated by the COVID-19 pandemic: livestock processing plants were among the top industries affected, comprising $6$ to $8\%$ of total cases and $3$ to $4\%$ of total deaths by mid July $2020$ \citep{taylor2020, bir2021the}. The labor shortages experienced in the meat industry can also be attributed to the public perception that meat industry jobs are often low-satisfaction careers \citep{Gastonandharrison2012} with increased risks of physical injury or post-traumatic stress disorder \citep{victor2016}. Today's meat industry needs safe, skilled, and precise workers that can maintain the rigorous standards necessary for animal welfare and food safety \citep{liang2021impact}. Overall, improving the availability of labor would support more streamlined and reliable supply chains, ensure consistent and rigorous animal welfare standards, and maintain reliability in meat product availability, quality, and price \citep{peel2021beef, wildmar2022perception, hobbs2021thecovid}. Progress toward a more consistent and reliable workforce would not only enable the industry to better meet public needs, but will also increase competitiveness in the face of international meat price volatility \citep{rzymski2021COVID-19}. 

% Front
\begin{figure*}[t!]
    \centering
    \includegraphics[width=2\columnwidth]{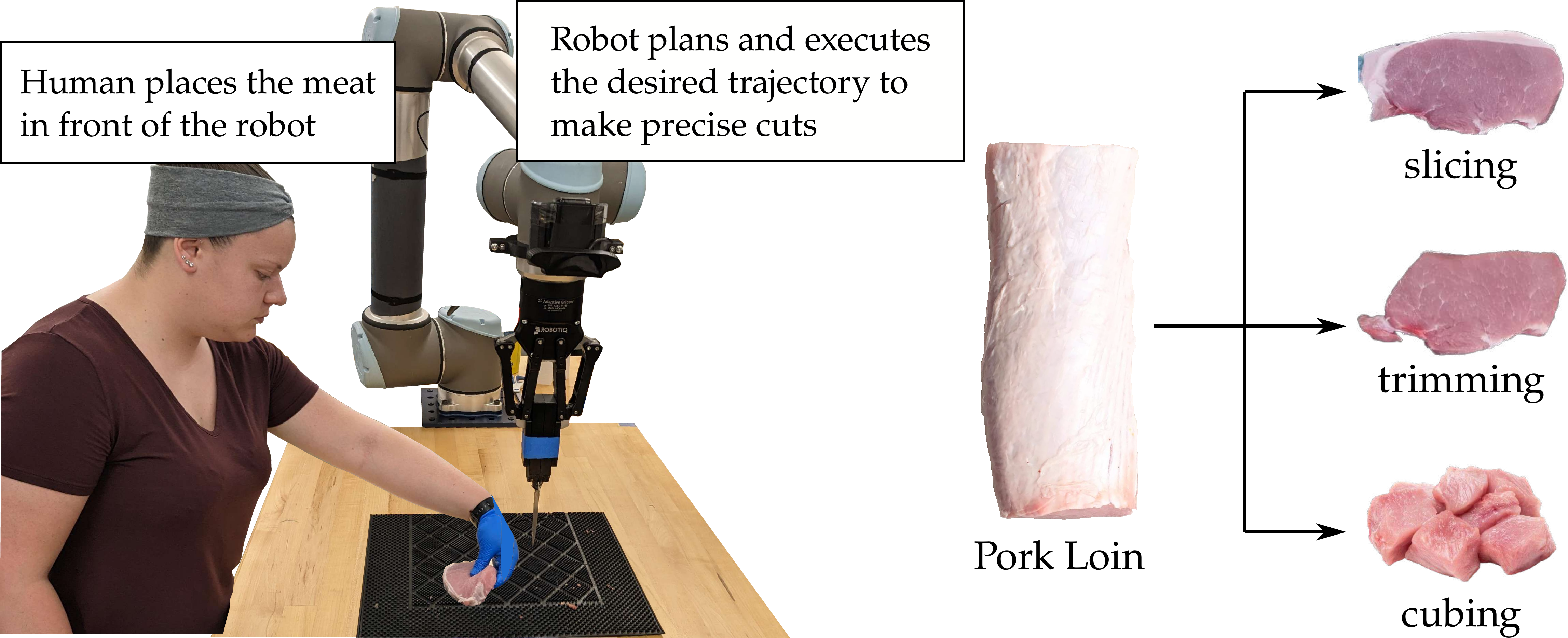}
    \caption{A multi-purpose robot arm for meat processing. (Left) A human collaborator places meat in front of the robot. Using an attached camera the robot detects the location of the meat. Under our proposed framework the robot can either plan and execute desired cuts that process the meat autonomously, or the robot can collaborate with the human to determine which cuts to make. (Right) In our experiments we leverage this framework to process a pork loin by slicing it into multiple cuts, trimming the fat from these cuts, and finally cubing the meat.}
    \label{fig:front}
\end{figure*}

Collectively, these labor supply challenges present an emerging opportunity for augmenting the workforce with robotic agents that assist human workers \citep{aly2023robotics, Madsen2006}. We recognize that there are already specialized machines that automate specific parts of the meat processing workflow (e.g., forming machines that shape $200,000$ nuggets per hour) \citep{hinrichsen2010manufacturing}, however these technologies are not flexible or collaborative. \textcolor{black}{For example, \citep{XIE2021106482,liu2017porcine} use 2D cameras for image segmentation and planning cuts, while \citep{metaRL} uses contact force feedback to adjust the cutting speed and cutting tool angle. These methods automate specific parts of the process where the robot operates in an isolated space. Similarly, the majority of existing technologies are built for a single purpose \citep{romanov2022towards} and traditionally replace human labor without opportunities for collaboration \citep{Barbut2020}.} To encourage widespread usage of automation in the meat industry, particularly for small and mid-sized plants, we must be able to account for carcass variability and dynamic workflows \citep{swenson2011exploring}. Looking beyond the meat industry, recent research in robotics has developed learning and control algorithms for multipurpose, collaborative robot arms. This includes control schemes to ensure robots are safe around humans \citep{lasota2017survey, haddadin2016physical, losey2018review}, as well as vision and learning tools that enable robots to adapt to new tasks \citep{osa2018algorithmic, mehta2022unified, parekh2022learning}. Recent growth of precision agricultural technologies and agricultural human-robot interactions is apparent in crop harvesting, fruit picking, and grading and sorting \citep{liu2020industry, klerkx2019review, marinoudi2019robotics}. The application of these same advances would directly address the needs for flexible and collaborative autonomous technologies within the U.S. meat industry; however, autonomous, collaborative, and multi-purpose meat processing robots have not yet been thoroughly investigated.

The objective of this work was to demonstrate a framework that enables a multipurpose robot arm in a shared human-robot space to perform example meat processing tasks (e.g., slicing, trimming, and cubing). After a human co-worker places the meat in front of the robot arm, our system autonomously detects where to cut, and then moves the robot arm (and attached knife) to complete these cuts. In this proof-of-concept demonstration of a flexible, collaborative approach to meat processing with robotic arms, we focus on two critical system requirements: safety and performance. \textcolor{black}{Safety concerns have been identified in human-robot collaboration due to the large force and rapid movements executed by robots in the workspace \citep{lasota2017survey}. Current human-robot collaborative technologies implement power and force limits, speed limits, as well as single action stopping function as mentioned in the technical specifications for collaborative robots (ISO/TS 15066). Further, they implement the use of collision avoidance strategies, most of which are based on distance sensing \citep{pedrocchi2013safe, flacco2012depth}. A broader suite of complementary sensors may be required in the application of collaborative technologies to meat processing, however, since contact is desirable between knife and meat but undesirable between knife and human collaborators.} We hypothesized that instrumentation, detection, and control systems to monitor robot location and knife contacts could be designed to support safety goals. Toward our performance objectives, we hypothesized that a combined vision and control approach could be designed to visually detect the meat, fat, and any markers placed by the human and enable precise cutting by mapping visual inputs to robot trajectories and subsequently executing those trajectories. We performed experiments with a $6$ degree-of-freedom multi-purpose robot arm, and tested these safety and performance frameworks during the completion of the example meat processing tasks.

%% file: method.tex
\section{Materials and Methods} \label{sec:method}

\begin{figure}
    \centering
    \includegraphics[width=\columnwidth]{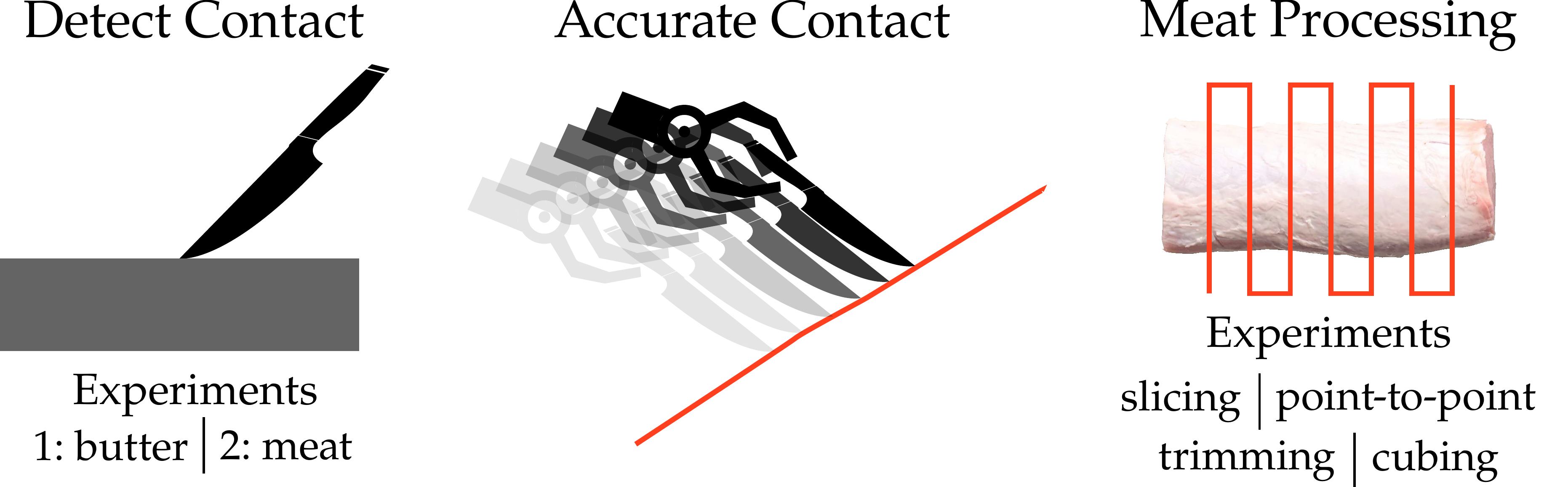}
    \caption{Left: safety precautions for avoiding human-robot collision. Here we constrained the robot's motion into a safe operating region above the cutting board. We also designed an instrumented knife for detecting unexpected contacts between the knife and another object. We conducted two experiments using the knife to cut butter and meat. We discuss the safety framework in Section \ref{sec:M1} and the results of the experiments in Section \ref{sec:RS}. Middle: testing the error between the robot's planned motion and the desired cut (i.e., the robot's cut precision). This is discussed in Section \ref{sec:M2}. Right: example meat processing using our vision and control framework. We enable the robot to autonomously detect meat and fat (vision) and then control its motion to cut the meat into various products. We test our approach on four meat cutting operations: slicing, removing fat autonomously (trimming) or in collaboration with humans (point-to-point), and cubing. More details of our vision and control framework are provided in Section \ref{sec:M2}. The results of the meat processing with this framework are listed in Section \ref{sec:RT}.}
    \label{fig:exp_summary}
\end{figure}

To progress towards our long-term objective of integrating multi-purpose robot arms into meat processing, we explore approaches to support safety and performance of a flexible, collaborative meat processing system built on $6$ degrees-of-freedom (DoF) robot arms. This initial execution of the system enables a multi-purpose robot arm to perform basic processing tasks including slicing a loin into chops, trimming excess fat, and cutting meat into cubes. While executing these tasks, we developed strategies that lay the groundwork to ensure the system and attached knife are safe around human workers, and that the system can accurately plan and execute the desired cuts either autonomously or in collaboration with humans. Here we discuss in detail our proposed framework and the experimental procedures used for deploying the groundwork of this framework. In Section \ref{sec:M1} we outline our bilateral approach for ensuring safety. In Section \ref{sec:M2} we discuss our physical setup and explain our vision and control algorithm for automating the robot arm to perform precise cuts. We outline the experiments in both sections in \fig{exp_summary}.

% Summary

\subsection{Safety: Preventing Unanticipated Collisions} \label{sec:M1}
In order for human coworkers to seamlessly and confidently collaborate with a meat processing robot arm, we first must ensure that the robot is safe. Consider the example in \fig{front} where a human is placing meat on the table in front of the robot: we need to prevent the robot from coming into contact with the human, and if an unexpected contact does occur, we need to detect that contact. Since the human and robot share a workspace and common objective, this is an instance of human-robot cooperation \citep{hri_categories}. ISO/TS 15066 is the technical specifications for collaborative robots which provides guidelines for risk assessment and designing collaborative operations. In accordance with these guidelines, the $6$ DoF robot arm must have speed and torque limits on the joints and must have a means to immediately stop the robot. Further, within physical human-robot cooperation, a major aspect of safety focuses on avoiding collisions between humans and robots \citep{DESANTIS2008253}. Intuitively, we can reduce the risk of collision by adding motion constraints for maintaining a safe distance between human and robot \citep{CHEMWENO2020104832}. Let the trajectory be the robot's motion path (i.e., the sequence of states the robot visits). To maintain a safe distance between this trajectory and the human, we will impose limits on the robot's workspace during task planning and execution. These limits constrain the robot into a safe region above the cutting board, so that nearby human workers know that the robot will always stay within this space. But even within this safe operating space, we need precautionary measures to detect when the robot collides with an object and to stop the robot's motion if this collision is unexpected or undesired \citep{DESANTIS2008253}. Accordingly, we demonstrate how an instrumented knife can be used to detect contact, and explore the data structures and approaches needed to sufficiently parameterize a system to enable robust contact detection through knife instrumentation.

\subsubsection{Constraints on the Robot}

% Safety
\begin{figure}[t]
    \centering
    \includegraphics[width=\columnwidth]{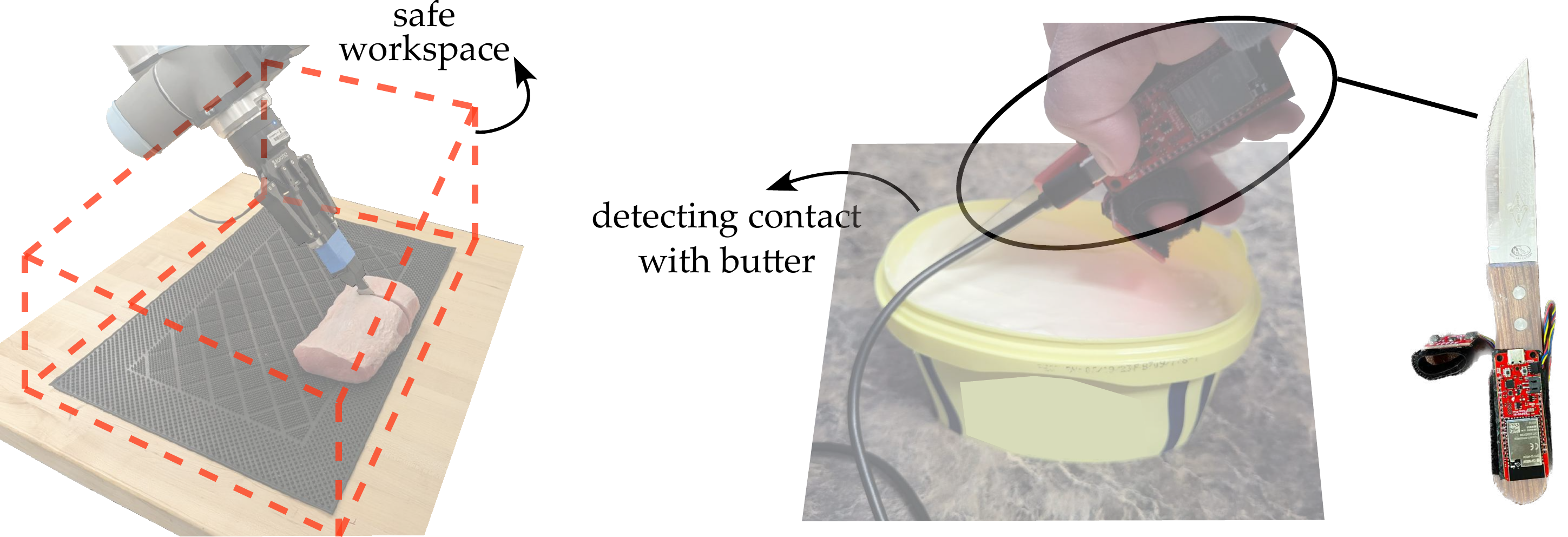}
    \caption{Schematic showing the robot's safety framework. The robot cuts the meat placed on a cutting board in front of it. (Left) We define a safe operating region around the cutting board, shown here in red, in which the robot is restricted throughout the operation. Staying within this bounded region helps to avoid undesirable contact with humans. (Right) We designed an instrumented knife for detecting undesirable contacts of the knife. This can inform the robot to re-plan its trajectory whenever needed to avoid collision with humans. To validate the effectiveness of the knife we perform two experiments: first, a human cuts butter with this knife; second, a human cuts meat with this knife. We test how accurately our setup can predict contact.}
    \label{fig:safety}
\end{figure}

At the start of the meat processing task the human collaborator places a piece of meat on the cutting board in front of the robot (see \fig{front}). We designate a fixed location on the table for meat placement. We then define a bounded region around this cutting board. In doing so, we constrain the robot's gripper and attached knife to always stay within the bounded region and avoid collision between the human and robot (see \fig{safety}). Mathematically, restricting robot movement is achieved by adding limits to the gripper's state. Let the gripper's state be $s^t = (x^t, y^t, z^t)$, where the superscript $t$ indicates the current timestep. We enforce limits on this state such that $x_{min} \leq x^t \leq x_{max}, \quad y_{min} \leq y^t \leq y_{max}, \quad z_{min} \leq z^t \leq z_{max}$. These limits are imposed both when the robot is planning a motion as well as when the robot is executing that motion. During planning, our framework calculates a desired trajectory. This trajectory is a sequence of robot states $\{(x^1, y^1, z^1), (x^2, y^2, z^2), (x^3, y^3, z^3), \dots \}$ that the robot must follow. After calculating this trajectory, each point within the trajectory is checked to ensure it lies within the safe region. Any point that is outside the safe region is adjusted back into the bounds of the safe region by setting the coordinate exceeding the region bounds as equal to those bounds. Specifically, if a point in the planned trajectory is $(1.1 x_{max}, 1.2 y_{max}, 1.05 z_{max})$, then this point is changed to $(x_{max}, y_{max}, z_{max})$. When the robot actually executes this planned motion, we implement a secondary check for safety. Specifically, as the robot progresses through each step along the motion plan it checks if the next step violates the workspace limits. If any step takes the gripper outside the safe region the robot does not execute that step. For instance, if the knife is at the boundary $(x_{max}, y_{max}, z_{max})$, any control action that would violate the limits is discarded to keep the knife within the limits. Collectively, these path-planning and path-execution checks effectively add constraints on the robot's motion to ensure that the knife always stays within the safe operating space. In theory, these constraints minimize the likelihood of the robot unintentionally interfering with any humans working in proximity to the robot. This control of the robot workspace is critical groundwork to support safety in collaborative human-robot environments \citep{DESANTIS2008253}. We note that this framework is flexible and can be modified to match the environment inside a meat processing factory; for example, designers can increase or decrease the dimensions and shape of this safe region to match available workspace or when moving the robot from one workstation to another.

\subsubsection{Experimental Setup of Instrumented Knife}

The constraints on the robot operating space reduce the risk of contact between human and robot outside the safe operation region in which the robot is restricted. During collaboration, however, the human can enter the robot's operating space, either intentionally (through actions like placing the meat on the table) or unintentionally. In these instances, we need additional safety precautions to prevent undesirable contacts between the human and the robot. Toward this need, we demonstrate how an instrumented knife can be used to detect contact, and use preliminary tests conducted while executing the example meat processing tasks to inform guidelines for more comprehensive safety protocol development. In this initial execution of the instrumented knife framework, we hypothesized that contact with meat could be determined by a proximity sensor and an inertial measurement unit (IMU). Further, we hypothesized that due to the unique nuances of the different cutting tasks, unique contact detection approaches would be needed for each task, but that these approaches could be generalized across individual pieces of meat. We collect data from a human using a sensor-equipped knife while performing three meat processing tasks (slicing, trimming, and cubing) to determine the accuracy of using the sensors for contact determination. Although beyond the scope of this experiment, determining that the knife is in contact with meat is a critical step toward a broader safety protocol which could use this contact detection along with visual and control inputs to explore whether the knife should be in contact with an object, and what kind of object the knife is in contact with. Here, we present the proof of concept that an instrumented knife can provide valuable feedback focused on contact detection for this eventual safety system.  

\p{Design of Instrumented Knife}
The knife was instrumented with a SparkFun ESP32-S2 Thing Plus (SparkFun Electronics, Niwot, CO) microprocessor, a SparkFun $9$ Degrees of Freedom (DoF) IMU Breakout, and a SparkFun Proximity Sensor Breakout using a hook and loop attachment for easy repositioning and cleaning. The IMU was selected as a candidate sensor because \citep{Wisanuvej2014blind} leveraged accelerometry as a tool for collision detection. Similarly, \citep{tsuji2020proximity} were successful in using time-of-flight proximity sensors for collision detection, thus we also selected proximity sensing as another candidate sensor. To accommodate ease-of-use and maximize logical placement of these sensors, the microprocessor and IMU were placed on either side of the knife handle, just below where the handle is gripped. The proximity sensor was placed below the handle, perpendicular to the knife, to allow it to aim along the knife blade toward the object being cut. These positions, as well as the selection of sensors were confirmed based on a preliminary test of reliability. In this test, the sensor readings were collected at $100$ hz while a human repeatedly cut into a block of butter. While the knife was in contact with the butter, the human pressed a button to code ground truth data representing contact. An initial test of the use of these sensor data to classify that contact yielded error rates <$1.1\%$ (data not shown). Based on the initial success of this instrumentation system confirmation exercise, we progressed to demonstrate the system in a meat processing context.  

\p{Data Preparation}
To determine whether this knife instrumentation system would be able to classify when the knife is in contact with meat, we conducted an experiment using the knife to perform the three target meat processing tasks (slicing, trimming, and cubing) on two pork loins. The experiment resulted in $23$, $26$, and $27$ replicates of slicing, trimming, and cubing. The differences in replicate number are due to some slicing actions not being recorded, differences in fat content among some slices (i.e., some slices did not need trimming), as well as the fewer number of slicing actions needed to create trimmable and cubeable slices. Based on success in the preliminary test with butter, the microprocessor controlling the instrumentation system was programmed using Arduino IDE to collect and log data from the proximity and IMU sensors at $100$ hz. This data collection resulted in $10$ features (i.e., independent variables) for use in training the contact detection algorithm. These features included the proximity reading, as well as the $x$, $y$, and $z$ axis readings of the accelerometer, magnetometer, and gyroscope. Ground truth measurements indicating when the knife was in contact with the meat were determined by the human operating the knife. When the human felt the knife come into contact with the meat, they pressed a button on the microprocessor. The button was continuously pressed during the entire time the knife was in contact with the meat. The microprocessor was coded such that this binary response variable (i.e., $1$ if pressed, $0$ otherwise) was logged with the $10$ associated sensor measurements. The average replicate resulted in $1,462$ observations, of which $69.5\%$ represented contact with the meat. The data were transferred in real-time from the microprocessor to local storage via universal serial bus (USB). Prior to analysis, each reading from each replicate was centered and standardized, and values exceeding $5$ standard deviations of the mean were omitted from analysis as presumed sensor errors.  

Because the ground truth observations were determined by when the human pressed the button, there was opportunity for human error. To minimize this, we visually confirmed that the human did not accidentally let go of the button during the cutting action by evaluating the consistency and duration of the indicator for contact in each cutting action. There will be residual human error associated with imperfect identification of the exact millisecond when the knife came into or exited contact with the meat; however, for the purposes of this proof-of-concept exercise, that error in ground truth coding was deemed acceptable. In future work exploring the refinement of this system for use in a broader safety protocol, high-speed imagery will be needed to confirm ground truth more precisely.  

\p{Data Analysis}
To explore the accuracy with which this prototype knife instrumentation system could be used to classify whether the knife was in contact with or approaching the meat, we trained a random forest classification algorithm (RF) using the randomForest package \citep{Liaw2002classification} of R v $4.2.1$ (R Core Team, $2022$). The target response to be classified was the binary indicator representing contact with the meat, and the features or independent variables used by the RF were the $10$ sensor readings. The RF is a supervised machine learning algorithm that is used to classify data by bootstrapping samples from the original data, building decision tees for each sample, and averaging the predictions from those trees in an ensemble to generate a final estimated outcome. The RF tends to be more robust than other classification approaches, with simple hyperparameter tuning and high prediction accuracy \citep{boateng2020basic}. To derive our RF, we split the data from each cutting task into $2$ subsets, with $60\%$ of the observations used for training, and $40\%$ used for independent evaluation of classification accuracy. The $60\%$ used for training was also used to tune the model parameters using the tuneRF function of the randomForest package. Based on this tuning, we bootstrapped $500$ samples from the training dataset, building $500$ trees with $4$ to $6$ variables tried at each split. The resulting tuned RF was then evaluated against the $40\%$ of held-out data to determine the number of true and false positives and negatives, as well as the overall error rate. The error rate was calculated as the number of false positives and false negatives divided by the total number of observations.  

To better understand the generalizability of the knife instrumentation system, we explored this training and testing strategy applied to three different data structures. In the first data structure (Superficial, within cut type; SWT), we combined all data from individual replicates within a cutting task into a single dataset for each cutting task. These data were then split $60/40$ as described above to benchmark the accuracy of the system when an individual algorithm is trained for each type of cut. This was a superficial split, meaning that all data were considered equally during the splitting into training and testing sets, without explicitly accounting for grouping factors like replicate. In the second data structure (Superficial, Across Types; SAT) we sought to explore how an algorithm could generalize across cut types. In this structure, we combined all data from the three cutting tasks, and split this combined data $60/40$ for training/testing, as described above. Again, this represented superficial splitting as replicates were not considered as a grouping factor when determining the data splits. In the third data structure (By Replicate, Within Type; RWT) we explored the impact of training and testing within individual cuts of meat. We trained the RF using the cut-specific datasets, but split such that $60\%$ of the replicates were used for training and $40\%$ of the replicates were used for testing. This meant that during testing, some replicates would reflect entirely ``new'' pieces of meat, which would be more representative of a real-world context where contact on new pieces of meat would need to be determined without prior opportunity to learn on data from that specific cut. Another series of random forest regressions was then applied to each of these data structures to evaluate the ability of the sensors to predict when an object was approaching. In this analysis, the same $60/40$ training/testing split was used to predict an incoming object in the $10$-$100$ milliseconds prior to contact.

In a real-world application of a full-scale safety control system, an acceptable error rate would be $0\%$. However, given that this instrumented knife is only one element of what could be incorporated into such a system, and we had potential for human error in pressing the contact button for determining ground truth in this proof-of-concept, we set a target error rate of $<3\%$. Time-of-flight proximity sensors such as the one employed in this experiment have accuracy and precision generally estimated as $1\%$ of the distance from the object. Based on the length of the instrumented knife, the expected precision was $1.5$ mm. The accuracy and precision of the IMU was expected to be $\pm0.5^{\circ}$ heading accuracy for the magnetometer, $1.5\%$ sensitivity for the gyroscope, and $0.5\%$ sensitivity for the accelerometer. Based on these hardware specifications, we expected that the random forest approach would support high fidelity detection of contact. 

\subsection{Performance: Planning and Executing Cuts} \label{sec:M2}

\begin{figure*}[ht!]
    \centering
    \includegraphics[width=2\columnwidth]{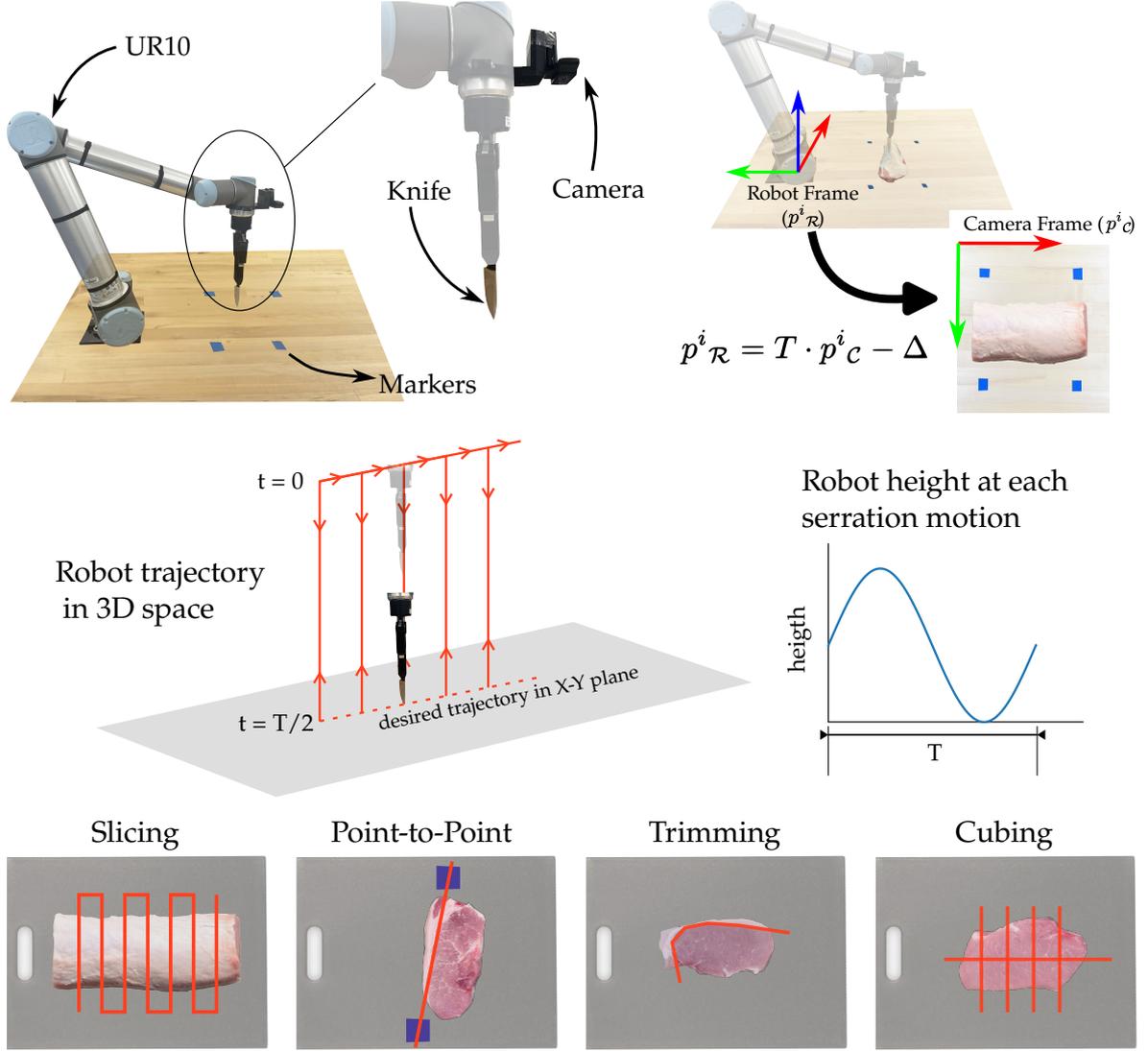}
    \caption{(Top) Our proposed robot vision and control framework. A camera is mounted on the robot arm. The gripper of the robot holds a knife for cutting. During calibration we use the position of the four  markers in the camera frame and the robot frame to optimize \eq{M1} and find the optimal parameters $\theta^*$.  (Middle) Robot motion in three dimensions. The output of the vision module is a trajectory in $x$-$y$ coordinates. While following this trajectory, the robot pauses at intervals and moves in the $z$ direction according to a sinusoid of time period $T$. This up-and-down motion moves the robot in and out of the meat to perform cuts. (Bottom) Example meat processing tasks. We design experiments with three tasks that are representative of meat processing applications in industry. First, meat is sliced into thin strips. Then, any excess fat on the strips is cut off; this is done through interaction with human coworkers (point-to-point) or completely autonomously (trimming). Finally, we produce uniform cubes from the slices (cubing).}
    \label{fig:method_tasks}
\end{figure*}

In the previous section we discussed the first part of our framework, which ensures the robot arm and attached knife are safe around humans. Next, we discuss the second part of our framework that enables the robot to precisely process meat. Specifically, we demonstrated a system that can autonomously or collaboratively perform meat processing tasks like slicing a loin into chops, trimming the fat off the meat, and cutting meat into uniform cubes. To complete these cutting operations, the robot must know where to make the cuts (i.e., vision) and it must be able to move with precision while cutting (i.e., control). Accordingly, for each new and previously unseen piece of meat, we leverage the robot's vision to plan a cutting motion based on the task. Next, we design a control algorithm that accurately follows the motion plan to make precise cuts. We highlight that each piece of meat used in the experiments can have different shape, size, and coloration. Moreover, the fat distribution on the meat slices has significant irregularity. Our method is capable of working under these sources of variability. We test the accuracy of our vision and control framework, and then perform the meat processing tasks (i.e. slicing, trimming, cubing).

\subsubsection{Experimental Setup for the Robot Arm}

We performed our experiments with a multi-purpose UR10 robot arm (Universal Robots). This robot has $6$ degrees-of-freedom and is commercially used by manufacturing industries for tasks like assembly, sanding, and welding \citep{ur10}. This robot complies with the ISO 10218-1 regulations for collaborative robots as mentioned on the manufacturer website (\url{https://www.universal-robots.com/articles/ur/safety/safety-faq/}). We mounted one RGB webcam (Logitech C290) on the robot arm using a custom $3$D-printed attachment. This camera was placed near the robot's gripper, moved with the robot, and could see the cutting board and any meat products on that board. We also rigidly attached a Henckels Everpoint stainless steel cutting knife to the robot's grippers. In the robot's home position this knife pointed down towards the table (see \fig{method_tasks}, top-left). Although our system uses this specific combination of robot, camera, and knife, the vision and control approaches are designed to flexibly extend to other robot, camera, and knife combinations.

In order for the robot arm to use the mounted camera we needed to convert from the camera's frame (i.e., the camera's perspective) to the arm's frame (i.e., the robot's perspective). For a summary of this conversion method refer to the \fig{method_tasks} (top-right). The camera captures objects in pixel coordinates, and in order for the robot to reach that object, it needs to know what $x$-$y$ position corresponds to the given pixel. We leveraged constrained optimization to calibrate the robot and camera \citep{rao2019engineering}. First, we measured the position of four markers in the robot's local coordinate frame (i.e., positions in meters). We then sent the robot to its home position and used the camera to get the coordinates of those same four markers (i.e., positions in pixels). Let $p_\mathcal{R}^i$ be the position of the $i$-th marker in the robot's frame and let $p_\mathcal{C}^i$ be the position of that same marker in the camera frame. The relationship between these two positions is given as 
\begin{equation}
    p{^i}_{\mathcal{R}} = T \cdot p{^i}_{\mathcal{C}} - \Delta
    \label{eq:M0}
\end{equation}
where $\Delta$ is some offset and $T$ is the scaled transformation matrix. We solved for the scaled transformation matrix and the offset by minimizing the total error between $p_\mathcal{R}$ and $p_\mathcal{C}$ subject to the constraint that the basis of this transformation were orthogonal \citep{lynch2017modern}. Let $\theta = (\theta_0, \ldots, \theta_4)$ be the parameters of this transformation matrix and offset. The optimal parameters $\theta^*$ are found using
\begin{equation} \label{eq:M1}
    \begin{split}
    \theta^* = &\text{arg}\min_{\theta}~ \sum_{i = 1}^4 \Bigg \| \underbrace{\begin{bmatrix}
        \theta_3 \\ \theta_4
    \end{bmatrix}}_{\Delta} + p_\mathcal{R}^i \\
    &- \underbrace{\begin{bmatrix} \theta_1 \cdot \cos(\theta_0) & -\theta_2 \cdot \sin(\theta_0) \\ \theta_1 \cdot \sin(\theta_0) & \theta_2 \cdot cos(\theta_0) \end{bmatrix}}_{T}p_\mathcal{R}^i \Bigg \|
    \end{split}
\end{equation}
The result of this optimization is a calibrated transformation (with parameters $\theta^*$) that maps camera pixel coordinates to robot arm coordinates; for any point the camera sees, the robot knows the corresponding point on the table's $x$-$y$ plane. Hence, the robot can reach for the objects that it sees through the camera. This calibration process needs to be repeated when the robot's workspace changes (i.e., when the robot is placed on a new factory floor, the operators would need to perform this calibration once). However, our framework automates the key steps of the calibration process, and operators would only need to input the measured positions $p_\mathcal{R}^i$ for each marker. The remaining steps could be autonomously performed by a provided a code package for increased flexibility.

\subsubsection{Robot Vision and Control} \label{sec:RM}

\textbf{Vision.} The robotic system developed in this work includes several elements, enabling vision (camera), actuation (knife and robot arm), and control (computer). We collectedly refer to these collaborative elements as the robot, with the understanding that individual elements have unique functions which work seamlessly to support the overall robot performance. To support the vision function of the robot, the camera mounted on the robot arm is controlled by the computer to collect and send visual information which the computer then processes and uses to inform the desired trajectory for the arm. In this workflow, the human partner places the meat on the cutting board, and the robot collects an image of the cutting board and surrounding environment. The robot then segments this image to separate out the meat, fat, and any visual markers placed there by the human co-worker. To identify all these objects the robot relies on color: i.e., for detecting meat the robot was programmed with an upper and lower range of red RGB values. The largest continuous red contour is classified as the meat, the largest white contour is labeled as fat, and any purple squares are treated as human-provided markers. The black cutting board provides a uniform background and the robot ignored any pixels outside of this cutting board. One advantage of our color-based approach is that it is robust to the meat product's shape (i.e., the meat could be more circular, more rectangular, or entirely irregular). 

We use three meat cutting tasks to test our approach. First, our framework cuts a loin into chops (\textbf{slicing}). Then it removes excess fat from the chops either autonomously (\textbf{trimming}) or under guidance from a human (\textbf{point-to-point}). Finally, it cuts the meat into uniform cubes (\textbf{cubing}). \fig{method_tasks} (bottom) shows example cuts for these tasks. In slicing and cubing the robot cut along straight lines to make uniform pieces. During slicing, the vision module planned $N$ equally spaced lines between the left and right edges of the meat. In point-to-point, the robot cuts along a straight line between the two markers placed by a human. This allows human partners to intervene and direct the robot: the human coworker can place and move the markers to indicate where the robot should cut, and then leave the robot to autonomously complete this cut. Cubing was similar to slicing --- our framework planned $N$ equally spaced lines between the top and bottom of the meat (making cuts orthogonal to the slices). In trimming, however, we need to cut around irregularly shaped fat, so simply drawing a straight line between the fat and meat is insufficient. Instead, we found the points at the intersection of the fat and meat contours, and then fit a trajectory to these points. To autonomously identify a smooth and efficient trajectory (i.e., motion path) between the fat and the meat, we used the ``Spatial QUalIty Simplification Heuristic - Extended'' or SQUISH-E algorithm \citep{muckell2014compression}. SQUISH-E inputs an initial trajectory of all the points between the fat and the meat, and then compressed that trajectory to a minimum number of points while ensuring that the error between the compressed and the original trajectories was below a given threshold. The resultant vision framework detects the meat and outputs a desired planar trajectory (i.e., $x$-$y$ motion) to make the desired cuts in each task.

\p{Control} We next controlled the robot to accurately follow these desired trajectories and perform cuts. There were three aspects of this control problem: (a) orienting the knife so that the blade was in the direction of the desired cut, (b) precisely following the desired $x$-$y$ trajectory along the table, and (c) modulating the height of the knife to cut the meat products.

Let $s^t = (x^t, y^t, z^t, \phi^t)$ be the robot's actual state at time $t$. This state consists of the $x$, $y$, and $z$ position of the gripper and $\phi$, the orientation of the knife. Our control objective is to drive $s^t$ towards the desired state $s_d^t = (x_d^t, y_d^t, z_d^t, \phi_d^t)$. Here $x_d$ and $y_d$ are the desired planar position of the knife that are output by our vision module. To get the desired height $z_d$, we programmed the robot to follow a sinusoidal up-and-down motion along the $z$ axis. Each time the robot moved down, the blade of the knife cut into the meat. As the robot came up, the blade was removed from the meat. The robot either moved in the $z$ direction or in the $x-y$ plane. When moving in $x-y$ plane the robot stopped at equal intervals to move up-and-down sinusoidally with a cycle time $T$. After completing the movement in $z$ the robot resumed moving along the $x-y$ trajectory at a constant height $z_d$. A schematic of this motion is shown in \fig{method_tasks} (middle), where the robot's three dimensional trajectory is highlighted in red. Next, to get the correct orientation of the knife $\phi_d$ we calculated the angle between the desired position at time $t$ and the next desired position at time $t+1$. We applied velocity control to guide the robot along the desired motion. We define $u^t$ as the robot's velocity command at time $t$: we computed this velocity using proportional feedback $u = K(s_d^t - s_d)$, where $K$ is an overall control gain. Prior robotics research has shown that this controller is stable and causes the actual state $s^t$ to converge to the desired state $s_d^t$ \citep{spong2006robot}. We implemented our robot controller at a frequency of $1$ kHz. During implementation we also incorporated the safety measures described in Section \ref{sec:M1}: the robot was constrained to move in a region around the cutting board to ensure safety for the humans who interacted with and worked around the system.

\subsubsection{Robot Accuracy}
Before we use our vision and control framework to process meat, we first need to test the efficacy of the vision and control modules of our framework. Accordingly, we performed a preliminary experiment with three representative tasks: using point-to-point cuts to move from side-to-side (horizontal) and from front to back (vertical), and trimming the fat to move along a curve (trim). These horizontal, vertical, and trimming motions are the primitive movements that our robot needs when performing meat cutting operations. For each task we conduct three trials. During the experiments, we record the planned trajectory output by the vision module $\{(x^1_d, y^1_d, z^1_d), (x^2_d, y^2_d, z^2_d), (x^3_d, y^3_d, z^3_d), \dots \}$ as well as the robot's actual trajectory while moving $\{(x^1, y^1, z^1), (x^2, y^2, z^2), (x^3, y^3, z^3), \dots \}$. In order to be accurate, our method should enable the robot to perform the planned motion with minimum error, i.e., the distance between the planned and the actual trajectories should be minimal.

\subsubsection{Meat Cutting Procedure}
To test our proposed vision and control approach for safe and autonomous robot cutting we performed studies with multiple meat products. In these studies the proctors placed fresh pork loins on a table in front of the robot arm, and the robot had to safely and precisely process the meat. We divided the meat processing into three representative sequential tasks: slicing the loin into chops (\textbf{slicing}), separating the detected fat from the meat of each chop autonomously (\textbf{trimming}) or by interacting with humans (\textbf{point-to-point}), and cubing the chop (\textbf{cubing}). Although not intended to represent a high-value workflow for the meat processing industry, this sequence of cuts allowed proof-of-concept that the robot could execute diverse and sequential tasks on a piece of meat collaboratively with human partners. Please refer to \fig{method_tasks} (bottom) for examples of these tasks. Throughout this process we measured the robot’s trajectories and the accuracy of the cuts, and assessed whether the final products met the desired industry specifications.

\p{Meat Products} In this proof-of-concept demonstration, the robot processed four pork loins. Each cut of pork was approximately $1.696$ kg and contained a layer of fat surrounding the top side of the meat. The meat processing of the pork loins was done sequentially; first, the robot cut the pork loins into a total of $33$ chops. It then removed the excess fat from each chop before cutting each chop into uniform cubes, producing a total of $158$ cubes.

\p{Independent Variables} To identify and make cuts, the robot followed the vision and control framework described in Section~\ref{sec:RM}. We varied the robot’s task across three levels: slicing, removing fat, and cubing. When slicing, the robot attempted to cut the loin into equally sized chops; the chops numbered between $8$ and $10$ depending on the individual pork loin. For removing fat we compared two strategies: point-to-point (a collaborative approach) and trimming (a fully autonomous approach). The point-to-point strategy allowed humans to interact with the robot: we asked proctors to place two felt markers on either side of the meat product to cut away fat, and the robot cut a straight line between those markers. Put another way, this allowed proctors to guide the robot in making desired cuts. Our alternative strategy was autonomously trimming the fat: when trimming, the robot attempted to automatically cut the fat that it detected from the sides of the meat based on the visual inputs. Finally, when cubing, the robot made final cuts to produce equally sized squares of meat about $3 \times 3$ cm in dimensions. As the cubes are made from the cuts of meat after fat removal, the height of the cubes is determined by the thickness of the chop from which they are cut. Hence, we only include the width and length of the cubes in our analysis.

\p{Dependent Variables} We used both qualitative analysis and quantitative metrics to evaluate the robot’s performance. From the robot’s perspective, we recorded the robot’s desired trajectory with states $s_d$ and the robot’s actual trajectory with states $s$. We then calculated the error in the robot’s motion, where error was defined as: Error $ = \frac{1}{n}\sum_{t=0}^T \| s_d^t - s^t\|$. Here, $n$ is the number of discrete points in the robot trajectory. When focusing on the meat, we measured the consistency and accuracy of the robot’s cuts. Here, consistency includes the dimensions and weight of each piece of meat the robot produced: for the slicing and cubing tasks, the robot should output meat products of almost identical sizes and weights. Accuracy refers to the precision of the robot’s cuts when separating the meat from the fat. In both point-to-point and trimming tasks, we took the pieces that the robot had cut away from the main body of meat and measured the dimensions of the fat and meat layers removed. We also weighed the meat removed during trimming. Ideally the robot should only remove fat, and the size and weight of the removed meat should be close to zero.

Besides these objective measures, we also analyzed the performance of the robot subjectively in point-to-point and trimming. We took photos of the meat products before and after the robot had completed the cut. We then conducted an online survey with $7$ anonymous, independent experts recruited from a number of meat processing facilities across the country. These experts from companies such as Smithfield, Tyson, Cargill, etc. had an average work experience of $8$ years in the meat industry. In the survey we first informed the experts about the robot's objective of separating the fat and meat. We then showed an image of each meat slice before and after the robot removed the fat, and asked the experts to rate the robot's performance on a scale of $1-7$. We emphasize that the experts did not know which method --- point-to-point or trimming --- was used to remove the fat. We divided the scores into three categories to classify the performance. A score $\leq 2$ was labelled poor, a score in the range $3-5$ was labelled fair, and a score $\geq 6$ was labelled good. Each expert was asked to rate a total of $31$ slices. We excluded $2$ slices from the survey since they did not have enough fat for removal. We counted the expert ratings for each of the three categories and calculated the percentage of times the experts ranked the cuts as poor, fair, or good.

%% file: results_discussion.tex
\section{Results and Discussion} \label{sec:results}

% contact detection results
\begin{table*}[h!]
\centering
\begin{tabular}{ |p{6cm}|p{1.5cm}|p{1.5cm}|p{1.5cm}|p{1.5cm}|p{1.2cm}|  }
 \hline
Data & Action & False Positives & False Negatives & Total Occurrences & Error Rate\\
 \hline
Superficial, within cut type (SWT) & Slicing & 86 & 168 & 13,670 & 1.86\%\\
 & Trimming & 143 & 300 & 11,998 & 3.69\%\\
 & Cubing & 236 & 328 & 18,779 & 3.00\%\\
By replicate, within cut type (RWT) & Slicing & 30 & 304 & 8,123 & 4.11\%\\
 & Trimming & 147 & 79 & 11,948 & 1.89\%\\
 & Cubing & 826 & 779 & 16,769 & 9.57\%\\
Superficial, across types (SAT) & All Cuts & 802 & 1148 & 44,090 & 4.42\%\\
 & & & & &\\
SWT Approaching Contact & Slicing & 0 & 88 & 13,438 & 0.007\%\\
 & Trimming & 2 & 91 & 12,135 & 0.008\%\\
 & Cubing & 0 & 108 & 18,924 & 0.006\%\\
RWT Approaching Contact & Slicing & 0 & 87 & 8,123 & 0.011\%\\
 & Trimming & 0 & 17 & 12,135 & 0.001\%\\
 & Cubing & 0 & 133 & 16,769 & 0.008\%\\
SAT Approaching Contact & All Cuts & 5 & 305 & 44,448 & 0.007\%\\
 \hline
\end{tabular}
\caption{Results of random forest classifications from experiments with instrumented knife. This includes both detecting contact types (Top) and detecting if the knife is approaching contact (Bottom)}
\label{table:1}
\end{table*}

Our high-level goal is to enable multi-purpose robot arms to safely and autonomously process meat while collaborating with human co-workers. In Section~\ref{sec:method} we introduced a formalism for safety (bounding the robot's motion and instrumenting the knife to detect contact) and for autonomous meat processing (visually sensing the meat and then controlling the arm to precisely cut the meat into various products). In this section, we discuss the results of our experiments. First, in Section \ref{sec:RS} we summarize the results of our instrumented knife and identify future work needed to leverage the instrumentation system for a broader safety framework. Second, in Section \ref{sec:RT} we discuss the results for the autonomous meat cutting tasks, including slicing, trimming, point-to-point, and cubing motions. Videos of the meat processing tasks and our proposed framework can be found here: \url{https://youtu.be/56mdHjjYMVc}

\subsection{Safety through the Instrumented Knife System}\label{sec:RS}
Here we list the results of the experiments from Section~\ref{sec:M1} that test our instrumented knife protocol, a critical first step toward a comprehensive safety system. We first report how accurately the approach was able to detect contact while performing the desired cuts. Next, we explore how different approaches to algorithm detection inform criteria and focus for future data collection and system refinement efforts. Three specific analyses were compared, including superficial data splitting within cut task (SWT), data splitting by replicate within cut task (RWT), and superficial data splitting across cut tasks (SAT).

In the SWT analysis, the instrumented knife system was able to detect contact with the meat with error rates ranging from $1.86\%$ to $3.69\%$ (Table~\ref{table:1}). Although the error rate for trimming exceeded our desirable threshold, the error rates achieved for slicing and cubing positively reflect the capacity of the system to predict contacts, and support future use of this instrumentation in developing safety algorithms for collaborative meat processing. Our instrumentation system builds on the work of \citep{koyama2018high-speed} by advancing past the exclusive use of proximity as a contact detection tool. Similarly, we advanced the IMU-based approach advocated by \citep{Wisanuvej2014blind} to also include proximity. By bringing together these two sensor systems, we were able to generate high-fidelity detection of contact from limited and preliminary data set. Building on the testing of our instrumentation system to include more data and improve the precision of ground truth labeling is expected to further advance the utility of this knife instrumentation system within the context of a broader safety management system. Based on these data and their consistency with previous findings, we demonstrate that cut-type-specific algorithms can be developed to identify contacts between the instrumented knife and the meat. 

Before designing experiments to support the use of a knife instrumentation system in broader safety management protocols, it is critical to address methodological questions relevant to the training and refinement of the instrumentation system. A key performance indicator for a successful contact detection system is the ability to recognize contact in novel situations. The superficial data splitting approach used in the SWT analysis does not allow for exploration of how the model performs on previously “unseen” objects. Therefore, to address the question of whether the algorithms trained in this assessment were generalizable to new pieces of meat within each cut type, we adjusted our training and testing data to test against entirely new pieces of meat (RWT analysis). In the RWT analysis, error rates were acceptable for trimming ($1.89\%$) but excessive for slicing ($4.11\%$) and cubing ($9.57\%$; Table~\ref{table:1}). Because the average error rate from the RWT analysis was higher than in the SWT assessment, we can conclude that the system suffers from a lack of generalizability. Generalizability (i.e., effective performance across multiple tasks and settings) is commonly identified as an important target in robotic control literature \citep{devin2018deep}, and a critical goal in the meat processing context \citep{maithani2021exoscarne}. In tests on generalizability in other contexts, RF approaches show strong performance among various machine learning algorithms tested \citep{odriscoll2021comparison}. However, superficial data segmentation for training and testing is acknowledged as a methodological pitfall likely to overestimate model performance \citep{maleki2023generalizability}. In this case, the error rates achieved from RWT are more reliable indicators of the expected algorithm performance in future scenarios than the error rates achieved through SWT. The confirmation of these differences in presumed algorithm performance is critical because it informs the appropriate experimental unit necessary for future system refinement. Specifically, in future model refinement tasks, we must focus on replicating a large number of meat cuts to allow for sufficient cut replicates for model training. A large number of observations per cut is important but insufficient to overcome the role of individual cuts in supporting derivation of more generalizable algorithms. 

In addition to being able to generalize to new meat objects, it is also critical that a safety system is able to generalize across tasks. Specifically, we want to ensure the robot can detect contact irrespective of the task that is being conducted. Although training unique algorithms for each cut type supports understanding of a ``best case scenario'' in terms of system performance, it is the ultimate goal of this approach to develop a single algorithm for contact detection that would work across any cutting task the robot is asked to perform. Toward this goal, we explored how the error rates were affected by training a single model across all cut types compared with models specific to each cut type (SAT approach). In this SAT analysis, the error rate was $4.42\%$ (Table~\ref{table:1}). Again, the elevated error rate here compared with the SWT analysis suggests a challenge with working toward a more generalizable system. Cross-task generalization in machine learning strategies is an active area of research, with various strategies tested for reinforcement \citep{frommberger2011task} and unsupervised learning \citep{lin2022unsupervised}. In supervised learning contexts, meta-datasets are often used as a key strategy to enhance cross-task generalizability \citep{mishra2021cross}. Our work supports the idea that a more comprehensive dataset containing a broader array of cutting tasks will be important for the future refinement of this system. In addition to replication among many cuts of meat, replication across meat processing tasks is also critical.

% Robot accuracy
\begin{figure*}
    \centering
    \includegraphics[width=2\columnwidth]{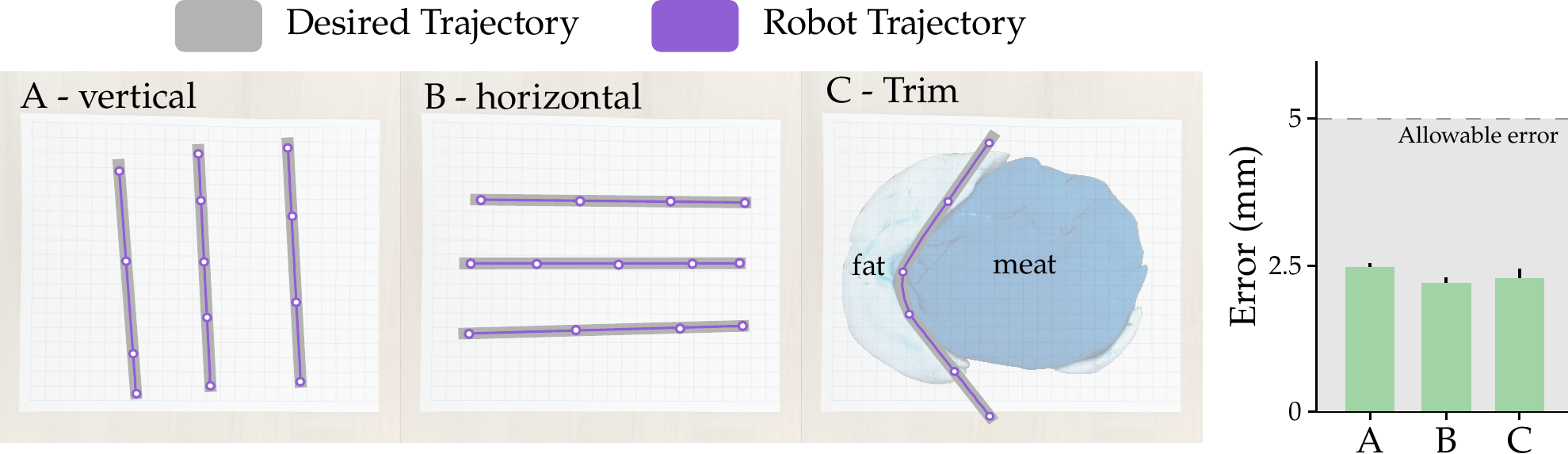}
    \caption{Results from our experiments testing vision and control accuracy on the robot arm. (Left) The three images show example trajectories for the three motions: vertical, horizontal, trimming. The trajectories planned by the vision module are shown in gray and the trajectories executed by the robot arm are highlighted in purple. (Right) Error between the desired trajectory and the robot's actual motion. The gray region shows the industry allowable threshold on Error (\citep{khodabandehloo2012robotics,nollet2006advanced}).}
    \label{fig:robot_accuracy}
\end{figure*}

Broadly, the majority of our testing showed error rates that exceeded the target level of precision. The few training and evaluation scenarios which yielded error rates below the target $3\%$ demonstrate the theoretical opportunity to move toward extremely accurate and precise contact detection using this instrumented knife approach. Toward the larger goal of ensuring that undesirable contact is prevented, we performed another series of random forest regressions using the same SWT, RWT, and SAT data structures. In these predictions, we tested the ability of the sensors to classify proximity to an object in the $10$ - $100$ milliseconds prior to contact rather than the moment of contact itself. This time range was selected based on the ability to stop robot movement prior to contact while still providing enough freedom for the human and robot to work collaboratively. Implementation of collision prevention models in other human-robot collaborations require $130$ milliseconds to calculate an incoming object \citep{shin2019real}. All error rates from the collision prevention models were less than $1\%$, which are notably lower than those for collision detection  (Table~\ref{table:1}). The collision prevention analysis using either SWT or RWT resulted in error rates between $0.006\%$ and $0.008\%$ or $0.001\%$ and $0.011\%$, respectively. Similarly, the SAT collision prevention analysis resulted in an error rate of $0.007\%$. The general lack of false positives is promising; however, the false negative rates indicate opportunities for future refinement. False negatives place a human collaborator at risk because they indicate the system would not be able to identify an impending collision. Given the ultimate goal of having a system capable of predicting object proximity with accuracy, the collision prevention approach using the SAT data was favorable for accurately detecting an incoming object in a timely manner, regardless of the task being performed. Both the error rates and the generalizability make incoming object detection the most promising of the tested approaches. We hypothesize that the lower accuracies of the contact detection models relate to the uncertainty surrounding the exact moment that contact is achieved rather than uncertainty of an incoming object. A major limitation of the collision prevention approach is, of course, that some ``collisions'' are desirable (i.e., the system needs to come into contact with meat). Future refinement to support the application of these approaches for automated meat processing should consider implementing a combination of collision prevention, object detection and differentiation (classifying objects that should be cut versus those that should not be), and contact determination for redundant and optimal safeguarding of humans working in collaboration with robots. Object detection and differentiation is outside the scope of the present project, but will be the goal of future work.

\subsection{Outcomes from Autonomous and Collaborative Meat Processing}\label{sec:RT}

Here we list the results of the experiments from Section~\ref{sec:M2} that test our vision and control framework for performing four meat processing tasks. We first report how accurately the robot was able to perform the desired cuts; i.e., how accurately the controller could execute the cuts output by the vision module. Next, we report the data from slicing, point-to-point, trimming, and cubing motions on four pork loins. This data includes objective measures of the work products and subjective evaluations by industry experts.

% Slicing
\begin{figure*}
    \centering
    \includegraphics[width=2\columnwidth]{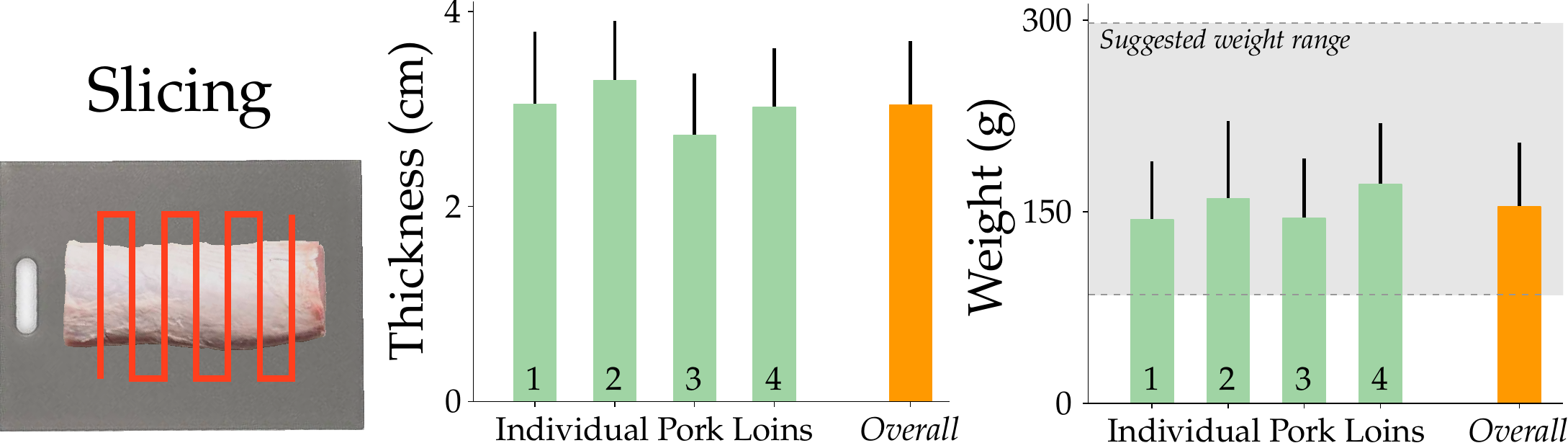}
    \caption{Results of slicing. (Left) An example trajectory planned by the robot's vision module. This trajectory makes equally spaced cuts from an overhead view of the meat. The number of cuts dependeded on the length of the pork loin. (Middle) We plot the \textit{Consistency} in the thickness of each slice. Green bars show the mean thickness of the slices produced from the same pork loin, i.e. pork loins $1$, $2$, $3$, and $4$. The orange bar shows the overall mean thickness of the slices across all pork loins. (Right) Plot of \textit{Consistency} in the weight of the slices. As before, the green bars show the average weight of the slices produced from the same pork loin. The orange plot shows the average weight of slices calculated across all four pork loins. The gray region shows the weight range suggested for pork chops by the USDA.}
    \label{fig:slicing}
\end{figure*}

\subsubsection{Accuracy of the Vision and Control Framework}
Before using our vision and control setup for meat cutting, we first tested our framework's ability to correctly plan and precisely execute desired cutting trajectories. The robot performed three different types of motions, namely, vertical, horizontal, and trim. Three trajectories of each type of motion are shown in \fig{robot_accuracy} (left). The gray lines are the trajectories planned by the vision module, and purple lines are the trajectories that were actually executed by the robot. We observe that the robot's trajectory coincides with the planned trajectory, indicating high accuracy in executing the desired motion. The plot on the right in \fig{robot_accuracy} quantifies the robot's error when performing these tasks. We see that the error, averaged across the three trials, is less than $2.5$ mm for each motion. The required accuracy in meat processing industries is generally $\pm 5$ mm \citep{khodabandehloo2012robotics,nollet2006advanced}. Hence, our robot controller causes the robot to accurately complete the cuts output by the vision module.

These results indicate that our method is adequately capable of meat processing applications. However, we recognise the drawbacks of detecting fat and meat based primarily on colors. Specifically, the downside of this mode of detection is that if the fat and meat have almost identical colors this visual method will fail, and we need to rely on additional modalities to distinguish between meat and fat. For example, the robot could palpate the meat to detect regions of differential stiffness \citep{ayvali2017utility, yan2021fast} or implement more comprehensive visual detection using near infrared spectroscopy \citep{zaid2020differentiation, falkovskaya2020literature}. This can be a new avenue for future research in the development of multi-purpose robot arms for meat processing.

\subsubsection{Autonomously and Collaboratively Processing Meat}

\textbf{Slicing.} In our test of the autonomous and collaborative meat processing, the proctor placed a pork loin on the cutting board and the robot autonomously sliced the pork loin into chops. In \fig{slicing} we report the consistency of those chops in terms of their size and weight. The image on the left shows an example trajectory the robot planned for slicing. The bar graphs show the robot's consistency in thickness (middle) and weight (right) for the slices. As a reminder, we wanted the robot to cut the meat into almost identical pieces that have similar thicknesses and weights. The average slice across all pork loins was $3.03$ cm thick and weighed $205.25$ g. The variance in slice thickness was $0.66$ cm or around $21.7$\% of the mean value, while the variance in weight was $66.85$ g or $32.57$\% about the mean. When we look at the slices for each individual pork loin, the weight showed a variance of $33.2$\%, $38.18$\%, $32.35$\%, $27.77$\% about the mean. The variance in the thickness for the pork loins was $24.67$\%, $18.5$\%, $23.53$\%, $20$\% respectively. Despite this variability, out of the $33$ total slices, $31$ were within the weight range suggested for pork chops in the Institutional Meat Purchase Specifications (\citep{specs}). Notably, the $2$ slices that do not meet the requirements were the first slices cut from two different pork loins. Since these slices were cut from the edges of pork loins that had irregular shapes, it was not unexpected that the weight and dimensions of these two slices did not meet the specifications.

One of the primary reasons why the strips had variable sizes and weights was that different pork loins have different thickness. Even for the same pork loin the thickness and width is far from uniform in terms of both the meat and fat present. Our method did not explicitly account for irregularities in the size of pork loins. Rather, the robot detected the meat and fit equally-spaced cuts across the meat. This one-dimensional approach to trajectory planning could be improved in the future by explicitly considering additional information about the meat and desired specifications for the cut. For example, utilizing the views from multiple camera angles can provide a much more accurate estimate of the meat's shape. Instead of simply fitting equally spaced lines across a overhead view of the meat, we envision future systems that build on our method to incorporate multiple camera angles to better divide the meat into equal portions. Such a system could eventually consider quality or value parameters, selecting the optimal breakdown for a primal based on visually sensed information and projected cut quality and value. In the meantime, this preliminary demonstration highlights that our proposed approach led to slices that satisfied the industry standard specifications.

\p{Removing Excess Fat} Once the pork loins were sliced, the next task was to remove any excess fat from each slice. In order to test how accurately the robot arm was able to remove the fat, we implemented two methods for comparison. The first method, point-to-point, was a collaborative approach where humans placed markers to guide the robot through each cut. In the second method, trimming, the robot performed the task in a fully autonomous manner using our developed algorithm.

% Removing fat
\begin{figure*}
    \centering
    \includegraphics[width=2\columnwidth]{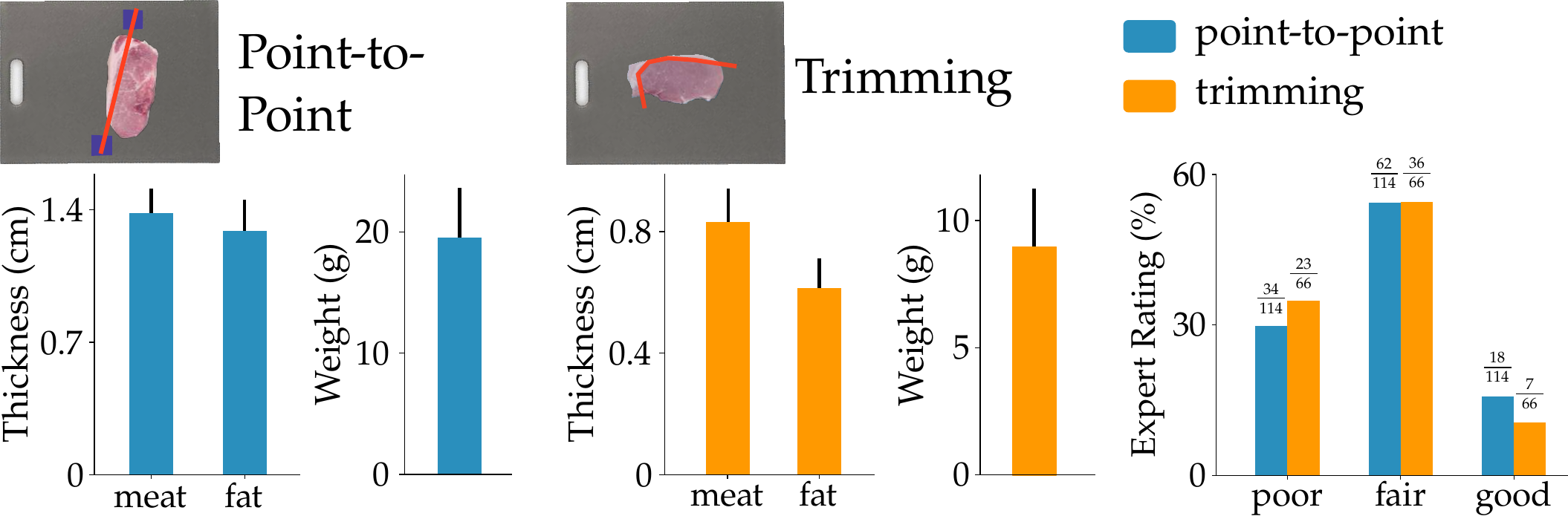}
    \caption{Results of point-to-point and trimming. (Left) Point-to-point: the robot cut between two felt markers placed by a human co-worker. The human was instructed to position the markers so that the robot's straight-line cut between the markers would separate fat and meat. The plot shows the mean thickness of the fat and meat removed during cutting as well as the mean weight of the meat removed. (Middle) Trimming: the robot cut a curve that was autonomously fit along the intersection of fat and meat. The plots show the average thickness of fat that was removed as well as the mean thickness and weight of the meat that was erroneously removed. (Right) Subjective results from our expert survey. The experts were shown images of the meat products before and after robot removed the fat using either point-to-point or trimming. Experts rated the robot's performance on a scale from $1-7$. The scores are classified intro three categories: a score of $1-2$ is poor, a score of $3-5$ is fair, and a score of $6-7$ is good. The frequency of each rating is calculated as a percentage of the total number of ratings. We compare the performance in point-to-point () with that in trimming (orange); the scores from both approaches indicate that the robot's cuts are fair or good roughly $70\%$ of the time.}
    \label{fig:removing_fat}
\end{figure*}

\p{Point-to-Point} In this task the proctors placed two markers near the meat indicating their desired cut, and the robot cut along a straight line between these markers. The image at the top of \fig{removing_fat} (left) shows an example cut made by the robot based on the proctor's guidance. The proctors used this approach (and placed their markers) to get the robot to cut fat away from the meat slices. The bar graphs at the bottom of this image show the average thickness of fat removed from each slice. We see that the robot was able to successfully remove an average of $1.28$ cm of fat with a variance of $0.68$ cm. Along with fat the robot also removed some amount of meat; the thickness and weight of this lost meat are also shown in these bar graphs. The thickness of meat removed averaged $1.38$ cm with a variance of $0.52$ cm and weight of the meat removed averaged $19.52$ g with a variance of $16.97$ g. We emphasize that with the point-to-point approach the robot is only cutting along straight lines between the human's markers.

% Qualitative comparison
\begin{figure}
    \centering
    \includegraphics[width=\columnwidth]{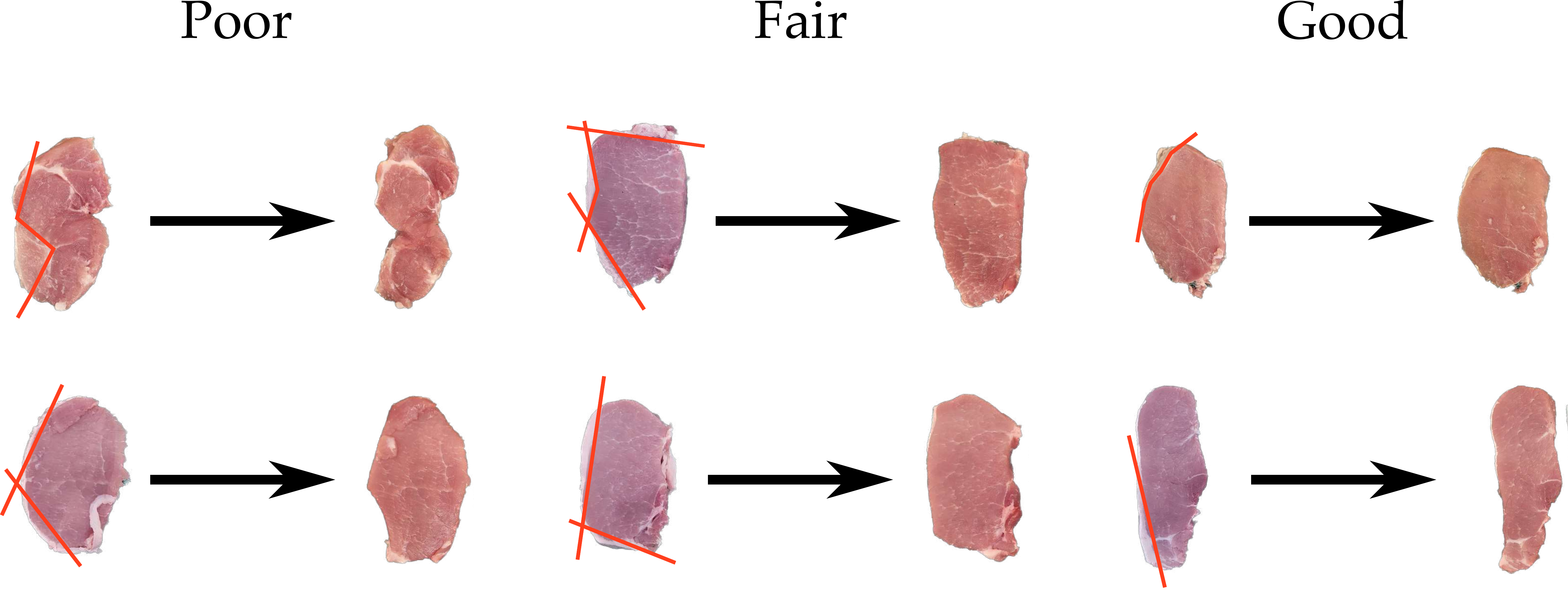}
    \caption{Examples showing the robot's performance in removing fat from the meat. We chose these examples based on the expert rating from our external survey. The lines in red show the trajectories the robot executed to cut away the fat. On the left we show the cuts that had the lowest expert rating. In these cuts the robot unintentionally removed a significant amount of meat along with the fat. The cuts in the middle were rated as fair (i.e., average) in the survey. In these the robot managed to remove all the fat along with a small quantity of meat. On the right we show the examples with the highest expert ratings. Here the robot removed all the fat while keeping the amount of meat lost to a minimum.}
    \label{fig:qualitative}
\end{figure}

\p{Trimming} During trimming the robot autonomously planned cuts to remove excess fat. It identified the fat and meat based on color, and made cuts at the intersection of the fat and meat. The image at the top of \fig{removing_fat} (middle) shows an example trajectory planned for trimming. The average thickness of fat removed in trimming was around $0.62$ cm with a variance of $0.44$ cm. The average weight of the meat that was also erroneously removed was $8.95$ g with a variance of $10.39$ g. The thickness of the removed meat had a mean of $0.83$ cm and a variance of $0.51$ cm. These results are plotted in \fig{removing_fat} (middle).

To understand if these cuts met industry standards, we conducted a survey with $7$ experts working in meat processing plants like Smithfield, Tyson, and Cargill. The experts rated each cut for point-to-point as well as trimming on a scale of $1-7$, where higher numbers indicated that the robot had performed a better cut. We divided this scale into poor ($1-2$), fair ($3-5$), and good ($6-7$). The plot in \fig{removing_fat} (right) shows the results from this survey. From the point-to-point images, $29.82$\% were rated poor, $54.38$\% were rated fair, and $15.78$\% were rated good. On the other hand, the chops trimmed through the fully autonomous system yielded $34.84$\% categorised as poor, $54.54$\% as fair, and $10.6$\% as good.

In \fig{qualitative} we show representative examples of the robot's performance in removing fat taken from point-to-point as well as trimming. We chose these examples based on the expert ratings of the cuts. We picked the two highest rated cuts, the two lowest rated cuts, and two cuts with a score close to the average. From these images we can see that the robot removed fat in every case; even for the slices that were rated as poor by the experts, the robot autonomously removed some fat. The difference in the quality of performance is largely due to the meat that was lost along with fat. Cuts that were scored more highly appeared to have less meat removed, and to maintain a more desirable fat layer around the shop.

The results of the survey revealed that experts preferred collaborative cuts executed via point-to-point over fully autonomous cuts performed using trimming. Under point-to-point the average score was $3.6$, and with trimming the average score was $3.48$. Since trimming cuts along a curve at the intersection of fat and meat while point-to-point cuts along a straight line, we originally expected that the more versatile trimming would perform better than point-to-point. In fact, our quantitative results confirm that trimming did indeed lead to a lower loss of meat while removing all the excess fat --- the average weight of meat removed by trimming ($8.95$ g) was less than half of that removed by point-to-point ($19.52$ g). But experts still slightly preferred point-to-point; this contrast in the objective and subjective results indicates that having a human-in-the-loop results in cuts that are more aligned with the expectations of the meat industry. We again highlight that experts were not told which method was used for cutting during the survey.

\begin{figure}
    \centering
    \includegraphics[width=\columnwidth]{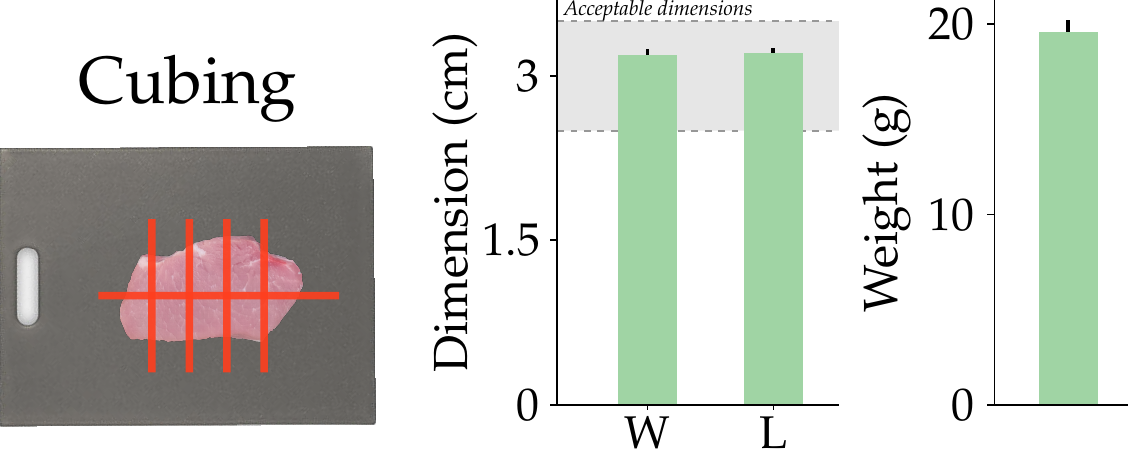}
    \caption{Results of the final meat processing task (cubing). After removing fat, the meat was diced into cubes of approximately the same size. The \textit{Consistency} in the length (L) and width (w) of the cubes are shown in the first plot. The shaded gray region outlines the acceptable range of variation in dimensions, which is $2.5-3.5$ cm according to (\citep{khodabandehloo2012robotics, nollet2006advanced}). The second plot shows the average weight of each cube. The vertical lines on each bar denote standard error.}
    \label{fig:cubing}
\end{figure}

\p{Cubing}
Finally, after removing the fat, the robot finally had to cut the slices of meat into equally sized squares of $3 \times 3$ cm. We measured the weight of each cube and found that the cubes weighed an average of $19.57$ g with a variance of $7.83$ g. The average dimensions (i.e., length and width) of the cubes were $3.31 \times 3.33$ cm respectively. We observed a variance of $0.51 \times 0.616$ cm in the length and width of the cubes respectively. For a summary of these results refer to \fig{cubing}. Overall, $11.4$\% of the total cubes fall within $\pm 0.3$ cm of our desired dimensions, and $48.1$\% of the total cubes lie within $\pm 0.6$ cm of our desired dimensions. We see a high variation in weight of the cubes even though the variation in the size of the cubes was not as high. In fact, half of the cubes fall within the $2.5 - 3.5$ cm range. This is the acceptable range given our specifications of making $3 \times 3$ cm squares and the accepted industry error of $\pm 0.5$ cm \citep{khodabandehloo2012robotics,nollet2006advanced}.

% Cubing

From these results we see that cubing suffers from the same drawbacks as slicing --- the meat slices have highly irregular shape and our method simply cuts along a grid of uniformly-spaced lines. We did not explicitly consider the shape of the meat or the desired dimensions of the final product when planning the cuts. In future works, the variability can be reduced by an informed planning process with multiple camera angles which takes into account the desired dimensions as well as the shape of the meat being cut. However, our current framework was successfully and autonomously able to produce $158$ cubes of meat, and the dimensions of these cubes fell within the industry accepted standards.

%% file: conclusion.tex
\section{Conclusion} \label{sec:conclusion}

In this work we developed and tested a framework that enables multi-purpose robot arms to perform multiple meat processing tasks in collaboration with human operators. Our framework included two parts: ensuring the system was safe, and enabling the robot to autonomously identify and cut meat. To ensure safety we first constrained the robot's motion into a designated region around the cutting board. We then developed a mechatronic knife that used IMU and proximity sensors to detect knife contact with meat. We trained algorithms to accurately perceive contact during representative meat processing tasks, and explored how data structured influenced the generalizability of these contact detection algorithms. Our results suggest that the instrumented knife is a promising strategy to identify knife contacts with meat products, but that future data collection efforts to improve generalizability of this framework are needed before it can be incorporated into a more comprehensive safety management system. These data collection efforts should focus on large numbers of replicated cuts and diverse cutting tasks. To ensure that contacts with objects other than meat can be identified, future data collection should also include replicated collisions with numerous and diverse non-meat objects. 

The second part of our framework combined vision and control to autonomously perform multiple meat processing tasks. We attached a camera to the robot arm, and calibrated the camera so that its measurements corresponded to the robot's workspace. We then developed a meat detection algorithm that used the camera to observe the location of the meat as well as the outline of any excess fat. Ultimately, the vision module output a trajectory of cuts that the robot should perform. We then developed a controller to make sure that the robot safely and accurately tracked these cuts. We specifically focused on processing pork loins by slicing them into chops, removing the fat from those chops, and then cubing the chops. During our experiments we enabled humans to collaborate with the robot by placing markers to guide the robot's cuts. Our results suggest that this vision and control approach leads to meat products that satisfy the industry standards in terms of size, weight, and removed fat, although considerable opportunity exists to improve system performance beyond this baseline. Industry experts rated around $70\%$ of the autonomous fat-removing cuts as either fair or good, and the dimensions of the meat products met the industry allowable margin of error. Overall, our results suggest that this novel framework can assist human workers in meat processing industries while meeting the industry's standards for safety and precision. Moving forward, our framework and results are a step towards technologies that help improve job satisfaction, supply chain consistency, and product availability within the meat industry.

\bigskip

\p{Acknowledgements}
\textbf{R. Wright}: Conceptualization, Data curation, Investigation, Methodology, Formal analysis, Writing - original draft.
\textbf{S. Parekh}: Conceptualization, Data curation, Investigation, Methodology, Formal analysis, Writing - original draft \& making figures.
\textbf{R. White}: Conceptualization, Supervision, Funding acquisition, Writing - review \& editing.
\textbf{D. Losey}: Conceptualization, Supervision, Funding acquisition, Writing - review \& editing.

% The roles are: Conceptualization; Data curation; Formal analysis; Funding acquisition; Investigation; Methodology; Project administration; Resources; Software; Supervision; Validation; Visualization; Roles/Writing - original draft; and Writing - review & editing. Note that not all roles may apply to every manuscript, and authors may have contributed through multiple roles.

\medskip

\p{Declaration of Competing Interest}
None. The authors declare no competing interests.

\medskip

\p{Funding}
This work was supported in part by the USDA National Institute of Food and Agriculture [grant number 2022-67021-37868].

\medskip

\p{Data Availability}
Data will be made available on request. Please contact either Ryan Wright (ryanw22@vt.edu) or Sagar Parekh (sagarp@vt.edu) to request this data.